\documentclass[11pt,onecolumn]{article}         
\usepackage[top=3.6cm, bottom=3.2cm, left=2.3cm, right=2.3cm]{geometry}

\usepackage{color}
\usepackage{graphicx}
\usepackage{cite}
\usepackage[cmex10]{amsmath}
\usepackage{amssymb}
\usepackage{subfigure}
\usepackage{rotating}
\usepackage{multirow}
\usepackage[colorlinks=true,linkcolor=red,citecolor=blue]{hyperref}
\usepackage{footmisc}
\newcommand{\ie}{\emph{i.e.}}

\begin{document}

\title{DOTA: A Large-scale Dataset for Object Detection in Aerial Images\footnote{DOTA dataset is available at \url{http://captain.whu.edu.cn/DOTAweb} or \url{https://captain-whu.github.io/DOTA}.}}

{
\author{Gui-Song~Xia$^{1}$\footnote{Equal contributions}, Xiang~Bai$^{2}$\footnotemark[2], Jian~Ding$^{1}$, Zhen Zhu$^2$, Serge Belongie$^3$, \\
Jiebo Luo$^4$, Mihai Datcu$^5$, Marcello Pelillo$^6$, Liangpei~Zhang$^1$\\
\\
$^1${\em State Key Lab. LIESMARS, Wuhan University, China}\\
{\tt \small \{guisong.xia, jding, zlp62\}@whu.edu.cn}\\
$^2${\em EIS, Huazhong Univ. Sci. and Tech., China}\\
{\tt \small \{xbai, zzhu\}@hust.edu.cn}\\
$^3${\em Computer Science Depart., Cornell University, USA}\\
{\tt \small sjb344@cornell.edu}\\
$^4${\em Computer Science Depart., Rochester University,  USA}\\
{\tt \small jiebo.luo@gmail.com} \\
$^5${\em German Aerospace Center (DLR), Germany}\\
{\tt \small mihai.datcu@dlr.de}\\
$^6${\em DAIS, University of Venice, Italy}\\
{\tt \small pelillo@dsi.unive.it}\\
}
}

\maketitle

\begin{abstract}
Object detection is an important and challenging problem in computer vision.
Although the past decade has witnessed major advances in object detection in natural scenes, such successes have been slow to aerial imagery, not only because of the huge variation in the scale, orientation and shape of the object instances on the earth's surface, but also due to the scarcity of well-annotated datasets of objects in aerial scenes.
To advance object detection research in Earth Vision, also known as Earth Observation and Remote Sensing,
we introduce a large-scale Dataset for Object deTection in Aerial images (DOTA).
To this end, we collect $2806$ aerial images from different sensors and platforms. Each image is of the size about $4000 \times 4000$ pixels and contains objects exhibiting a wide variety of scales, orientations, and shapes. These DOTA images are then annotated by experts in aerial image interpretation using $15$ common object categories. The fully annotated DOTA images contains $188,282$ instances, each of which is labeled by an arbitrary (8 d.o.f.) quadrilateral.
To build a baseline for object detection in Earth Vision, we evaluate state-of-the-art object detection algorithms on DOTA. Experiments demonstrate that DOTA well represents real Earth Vision applications and are quite challenging.
\end{abstract}
\definecolor{car}{RGB}{0,0,128}
\definecolor{parking lot}{RGB}{255,20,147}
\definecolor{plane}{RGB}{255,0,255}
\definecolor{baseball diamond}{RGB}{250,235,215}
\definecolor{bridge}{RGB}{0,128,0}
\definecolor{ground track field}{RGB}{127,255,212}
\definecolor{ship}{RGB}{165,42,42}
\definecolor{tennis court}{RGB}{138,43,226}
\definecolor{basketball court}{RGB}{222,184,135}
\definecolor{storage tank}{RGB}{95,158,160}
\definecolor{soccer ball field}{RGB}{127,255,0}
\definecolor{turntable}{RGB}{210,105,30}
\definecolor{harbor}{RGB}{255,255,0}
\definecolor{electric pole}{RGB}{100,149,237}
\definecolor{swimming pool}{RGB}{220,20,60}
\definecolor{lake}{RGB}{255,160,122}
\definecolor{helicopter}{RGB}{0,0,139}
\definecolor{airport}{RGB}{0,139,139}
\definecolor{overpass}{RGB}{184,134,11}
\definecolor{viaduct}{RGB}{184,134,11}
\definecolor{yellow}{RGB}{255,255,0}

\section{Introduction}

\begin{figure}[htp!]
\centering
\subfigure{
}
\includegraphics[width=0.9\linewidth]{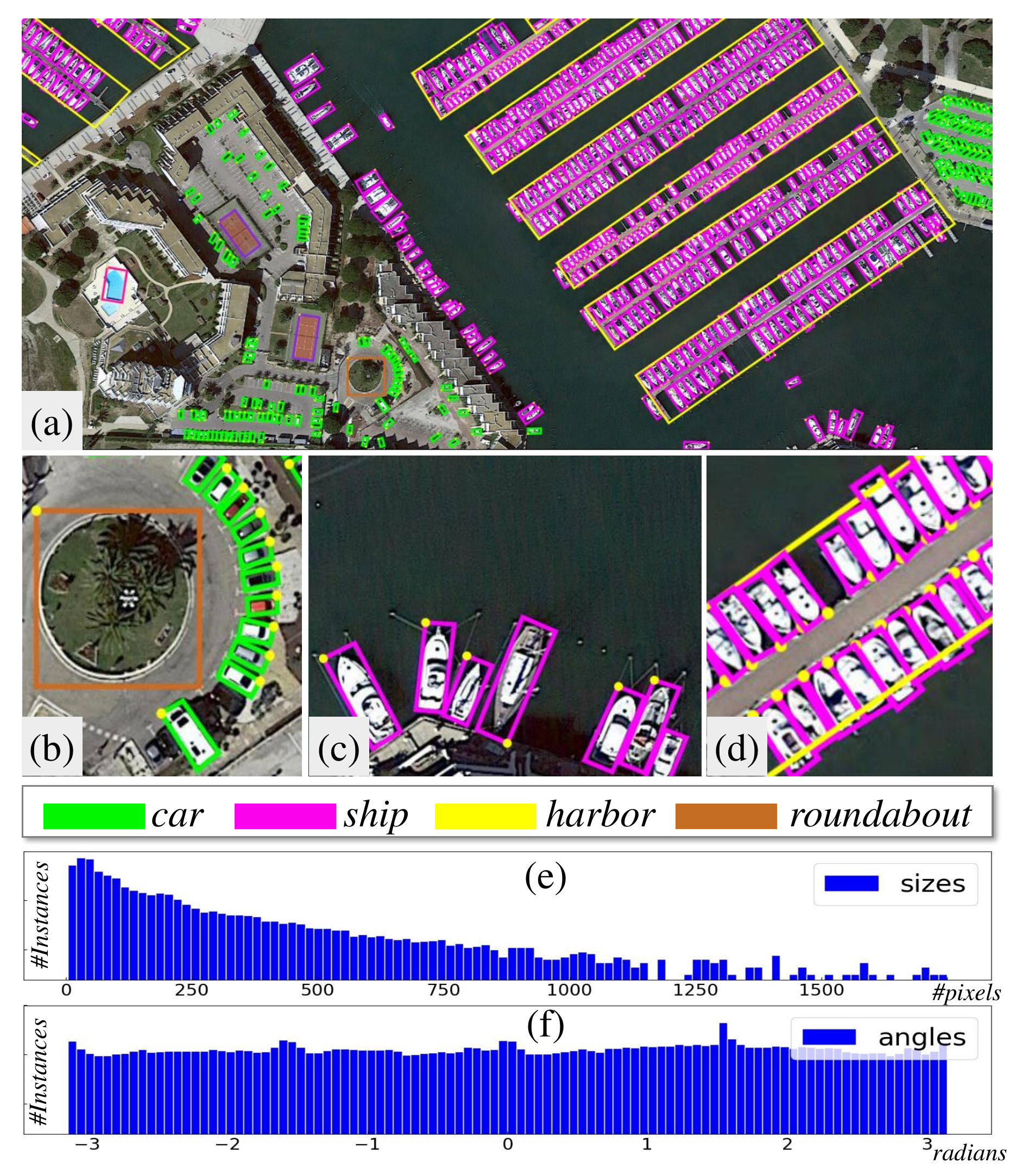}
\vspace{-2mm}
\caption{{\bf An example taken from DOTA}. 
(a) Typical image in DOTA consisting of many instances across multiple categories. (b) Illustration of the variety in instance orientation and size. 
(c),(d) Illustration of sparse instances and crowded instances, respectively. 
Here we show four out of fifteen of the possible categories in DOTA. Examples shown in (b),(c),(d) are cropped from source image (a). The histograms (e),(f) exhibit the distribution of instances with respect to size and orientation in DOTA.
}
\vspace{-2mm}
\label{fig:large-example}
\end{figure}

Object detection in Earth Vision refers to localizing objects of interest (e.g., vehicles, airplanes) on the earth's surface and predicting their categories.
In contrast to conventional object detection datasets, where objects are generally oriented upward due to gravity, the object instances in aerial images often appear with arbitrary orientations, as illustrated in Fig.~\ref{fig:large-example}, depending on the perspective of the Earth Vision platforms.

Extensive studies have been devoted to object detection in aerial images~\cite{LHI_method, rotation-invariant-radial-gradient,HRSC2016,roatation-invariant-cvpr,Detect_car_in_uav, plane_weakly_deeplearning,long2017accurate,rotation-invariant,wan2017affine,building1}, drawing upon recent advances in Computer Vision and accounting for the high demands of Earth Vision applications. Most of these methods~\cite{plane_weakly_deeplearning,long2017accurate,rotation-invariant,roatation-invariant-cvpr} attempt to transfer object detection algorithms developed for natural scenes to the aerial image domain.
Recently, driven by the successes of deep learning-based algorithms for object detection, Earth Vision researchers have pursued approaches based on fine-tuning networks pre-trained on large-scale image datasets (e.g., ImageNet~\cite{Imagenet} and MSCOCO~\cite{COCO}) for detection in the aerial domain, see e.g.~\cite{long2017accurate,buiding_detction_deeplearning1,VHR,roatation-invariant-cvpr}. 

While such fine-tuning based approaches are a reasonable avenue to explore, images such as Fig.~\ref{fig:large-example} reveals that the task of object detection in aerial images is distinguished from the conventional object detection task in the following respects:
\vspace{-2mm}
\begin{itemize}
\item[-] The scale variations of object instances in aerial images are huge.
This is not only because of the spatial resolutions of sensors, but also due to the size variations inside the same object category. 
\vspace{-2mm}
\item[-] Many small object instances are  crowded in aerial images, for example, the ships in a harbor and the vehicles in a parking lot, as illustrated in Fig.~\ref{fig:large-example}. Moreover, the frequencies of instances in aerial images are unbalanced, for example, some small-size (e.g. $1k \times 1k$) images contain $1900$ instances, while some large-size images (e.g. $4k \times 4k$) may contain only a handfull of small instances.
\vspace{-2mm}
\item[-] Objects in aerial images often appear in  arbitrary orientations.
There are also some instances with an extremely {large} aspect ratio, such as 
a bridge.
\vspace{-1mm}
\end{itemize}

Besides these distinct difficulties, the studies of object detection in Earth Vision are also challenged by the well-known dataset bias problem~\cite{torralba2011unbiased}, \ie the degree of generalizability across datasets is often low. In order to alleviate such biases, the dataset should be annotated to reflect the demands of real world applications.

Therefore, it is not surprising that the object detectors learned from natural images are not suitable for aerial images. However, existing annotated datasets for object detection in aerial images, such as UCAS-AOD~\cite{ucas-aod} and NWPU VHR-10~\cite{VHR}, tend to use images in ideal conditions (clear backgrounds and without densely distributed instances), which cannot adequately reflect the problem complexity.

To advance the object detection research in Earth Vision, this paper introduces a large-scale Dataset for Object deTection in Aerial images (DOTA).
We collect $2806$ aerial images from different sensors and platforms with crowdsourcing. Each image is of the size about $4000 \times 4000$ pixels and contains objects of different scales, orientations and shapes. These DOTA images are annotated by experts in aerial image interpretation, with respect to $15$ common object categories. The fully annotated DOTA dataset contains 188,282 instances, each of which is labeled by an oriented bounding box, instead of an axis-aligned one, as is typically used for object annotation in natural scenes.
%
The main contributions of this work are:
\begin{itemize}
\vspace{-2mm}
\item[-] To our knowledge, DOTA is the largest annotated object dataset with a wide variety of categories
in Earth Vision. It can be used to develop and evaluate object detectors in aerial images. We will continue to update DOTA, to grow in size and scope and to reflect evolving real world conditions.
\vspace{-2mm}
\item[-] We also benchmark state-of-the-art object detection algorithms on DOTA, which can be used as the baseline for future algorithm development.
\end{itemize}

\vspace{-1mm}
In addition to advancing object detection studies in Earth Vision, DOTA will also pose interesting algorithmic questions to conventional object detection in computer vision.

\section{Motivations}

Datasets have played an important role in data-driven research in recent years~\cite{LHI_CVPR,Imagenet,COCO,Places,EmotionRecognition}. Large datasets like MSCOCO~\cite{COCO} are instrumental in promoting object detection and image captioning research. 
When it comes to the classification task and scene recognition task, the same is true for ImageNet~\cite{Imagenet} and Places~\cite{Places}, respectively. 

However, in aerial object detection, a dataset resembling MSCOCO and ImageNet both in terms of image number and detailed annotations has been missing, which becomes one of the main obstacles to the research in Earth Vision, especially for developing deep learning-based algorithms. 
Aerial object detection is extremely helpful for vehicle counting, remote object tracking and unmanned driving. Therefore, a large-scale and challenging aerial object detection benchmark, being as close as possible to real-world applications, is imperative for promoting research in this field.

We argue that a good aerial image dataset should possess four properties, namely, 1) a large number of images, 2) many instances per categories, 3) properly oriented object annotation, and 4) many different classes of objects, which make it approach to real-world applications. 
However, existing aerial image datasets~\cite{ucas-aod,HRSC2016,DLR3KMunichVehicle,VEDAI} share in common several shortcomings: insufficient data and classes, lack of detailed annotations, as well as low image resolution. Moreover, their complexity is inadequate to be considered as a reflection of the real world.

\begin{table*}[htb!]
\small
\centering
\begin{tabular}{c|ccrrrr}
\hline
Dataset & Annotation way & \#main categories & \#Instances & \#Images & Image width \\
\hline
 NWPU VHR-10 ~\cite{VHR}  & horizontal BB & 10 & 3651 & 800 & $\sim$1000 \\
SZTAKI-INRIA ~\cite{SZTAKI-INRIA} & oriented BB & 1  & 665 & 9 & $\sim$800  \\
TAS ~\cite{TAS} & horizontal BB & 1 & 1319 & 30 & 792  \\
COWC ~\cite{COWC} & one dot & 1 & 32716 & 53  & 2000$\sim$19,000 \\
VEDAI ~\cite{VEDAI} & oriented BB & 3 & 2950 & 1268 & 512, 1024 \\
UCAS-AOD ~\cite{ucas-aod} & oriented BB & 2 & 14,596 & 1510 & $\sim$1000 \\
HRSC2016 ~\cite{HRSC2016} & oriented BB & 1 & 2976 & 1061 & $\sim$1100 \\
3K Vehicle Detection ~\cite{DLR3KMunichVehicle} & oriented BB & 2 & 14,235 & 20 & 5616 \\
\hline
\bf{DOTA} & oriented BB & \bf{14} & \bf{188,282} & \bf{2806} &  800$\sim$4000 \\
\hline
\end{tabular}
\vspace{-2mm}
\caption{Comparison among DOTA and object detection datasets in aerial images. BB is short for bounding box. {\em One-dot} refers to annotations with only the center coordinates of an instance provided. Fine-grained categories are not taken into account. For example, DOTA consist of 15 different categories but only 14 main categories, because small vehicle and large vehicle are both sub-categories of vehicle.}
\label{former-datasets}
\vspace{-2mm}
\end{table*}

Datasets like TAS~\cite{TAS}, VEDAI~\cite{VEDAI}, COWC~\cite{COWC} and DLR 3K Munich Vehicle~\cite{DLR3KMunichVehicle} only focus on vehicles. UCAS-AOD~\cite{ucas-aod} contains vehicles and planes while HRSC2016~\cite{HRSC2016} only contains ships even though fine-grained category information are given. All these datasets are short in the number of classes, which restricts their applicabilities to complicated scenes. In contrast, NWPU VHR-10~\cite{VHR} is composed of ten different classes of objects while its total number of instances is only around $3000$. Detailed comparisons of these existing datasets are shown in Tab.~\ref{former-datasets}. Compared to these aerial datasets, as we shall see in Section~\ref{statistics_DOTA}, DOTA is challenging for its tremendous object instances, arbitrary but well-distributed orientations, various categories and complicated aerial scenes. Moreover, scenes in DOTA is in coincidence with natural scenes, so DOTA is more helpful for real-world applications.

When it comes to general objects datasets, ImageNet and MSCOCO are favored by researchers due to the large number of images, many categories and detailed annotations. ImageNet has the largest number of images among all object detection datasets. However,  the average number of instances per image is far smaller than MSCOCO and our DOTA, plus the limitations of its clean backgrounds and carefully selected scenes. Images in DOTA contain an extremely large number of object instances, some of which have more than 1,000 instances. PASCAL VOC Dataset~\cite{PASCALVOC} is similar to ImageNet in instances per image and scenes but the inadequate number of images makes it unsuitable to handle most  detection needs.
%
Our DOTA resembles MSCOCO in terms of the instance numbers and scene types, but DOTA's categories are not as many as MSCOCO because objects which can be seen clearly in aerial images are quite limited.

Besides, what makes DOTA unique among the above mentioned large-scale general object detection benchmarks is that the objects in DOTA are annotated with properly {\em oriented bounding boxes} ({\textbf{OBB} for short). 
OBB can better enclose the objects and differentiate crowded objects from each other. The benefits of annotating objects in aerial images with OBB are further described in Section~\ref{Annotation-method}. We draw a comparison among DOTA, PASCAL VOC, ImageNet and MSCOCO to show the differences in Tab.~\ref{generalobjectscomparison}. 


\begin{table}[ht!]
\footnotesize
\centering
\begin{tabular}{ccccc}
\hline
Dataset                                     & Category 
& \begin{tabular}[c]{@{}c@{}}Image \\ quantity\end{tabular} 
& \begin{tabular}[c]{@{}c@{}}BBox \\ quantity\end{tabular} 
& \begin{tabular}[c]{@{}c@{}}\ Avg. BBox \\quantity\end{tabular} \\ 
\hline
\begin{tabular}[c]{@{}c@{}}PASCAL VOC \\ (07++12)\end{tabular}  & 20  & 21,503       & 62,199 & 2.89 \\ \hline
\begin{tabular}[c]{@{}c@{}}MSCOCO \\ (2014 trainval)\end{tabular} & 80 & 123,287 & 886,266 & 7.19 \\ \hline
\begin{tabular}[c]{@{}c@{}}ImageNet\\ (2017train)\end{tabular} & 200 & 349,319   & 478,806 & 1.37 \\ \hline
\textbf{DOTA} & 15 & 2,806 & 188,282         & \textbf{67.10} \\
\hline
\end{tabular}
\caption{Comparison among DOTA and other general object detection datasets. BBox is short for bounding boxes, {\em Avg. BBox quantity} indicates average bounding box quantity per image. Note that for the average number of instances per image, DOTA surpasses other datasets hugely.}
\label{generalobjectscomparison}
\end{table}


\section{Annotation of DOTA}
\label{Annotation-method}

\subsection{Images collection}
{As mentioned in ~\cite{earthvisonba}, in aerial images, the resolution and variety of sensors being used are factors to produce dataset biases. To eliminate the biases, images in our dataset are collected from multiple sensors and platforms (e.g. Google Earth) with multiple resolutions. 
To increase the diversity of data, we collect images shot in multiple cities carefully chosen by experts in aerial image interpretation. 
We record the exact geographical coordinates of the location and capture time of each image to ensure there are no duplicate images.} 



\subsection{Category selection}
{Fifteen categories are chosen and annotated in our DOTA dataset, including {\em plane, ship, storage tank, baseball diamond, tennis court, basketball court, ground track field, harbor, bridge, large vehicle, small vehicle, helicopter, roundabout, soccer ball field and basketball court}.

{The categories are selected by experts in aerial image interpretation according to whether a kind of objects is common and its value for real-world applications. The first 10 categories are common in the existing datasets,} e.g., ~\cite{DLR3KMunichVehicle,VHR,ucas-aod,COWC}}, 
{We keep them all except that we further split vehicle into large ones and small ones because there is obvious difference between these two sub-categories in aerial images. Others are added mainly from the values in real applications. For example, we select helicopter  considering that moving objects are of significant importance in aerial images. Roundabout is chosen because it plays an important role in roadway analysis.}

{It is worth discussing whether to take ``stuff" categories into account. There are usually no clear definitions for the "stuff" categories (e.g. {\em harbor, airport, parking lot}), as is shown in the SUN dataset ~\cite{SUN}. However, the context information provided by them may be helpful for detection.
We only adopt the harbor category because its border is relatively easy to define and there are abundant harbor instances in our image sources.}
 The final extended category is soccer field.
 
 In Fig.\ref{fig:category}, we compare the categories of DOTA with NWPU VHR-10~\cite{VHR}, which has the largest number of categories in previous aerial object detection datasets. Note that DOTA surpass NWPU VHR-10 not only in category numbers, but also the number of instances per category.

\begin{figure}[htb!]
\vspace{-2mm}
\begin{center}
\includegraphics[width=0.85\linewidth]{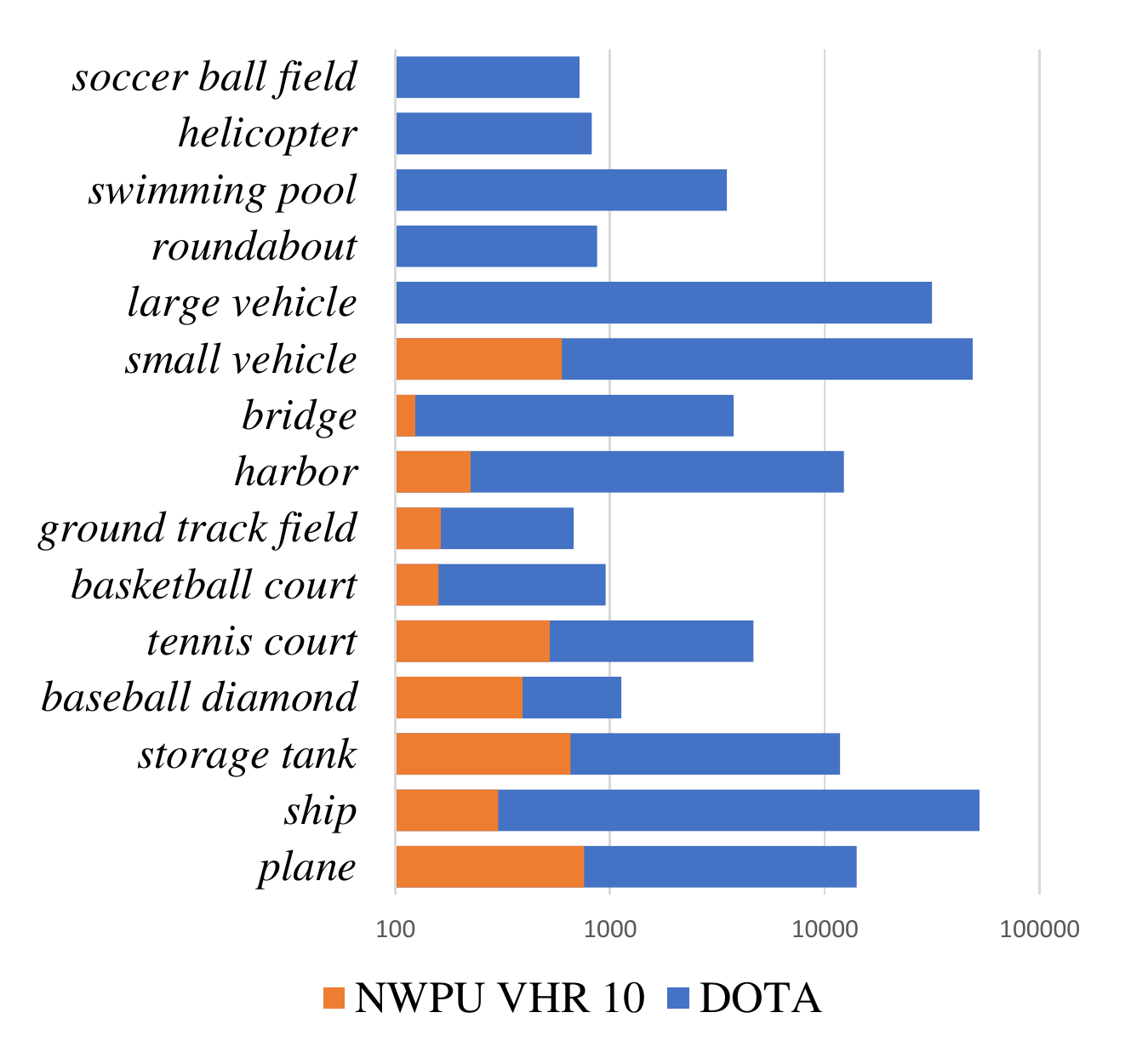}
\end{center}
\vspace{-2mm}
\caption{Comparison between DOTA and NWPU VHR-10 in categories and responding quantity of instances.}
\label{fig:category}
\end{figure}

\subsection{Annotation method}
{We consider different ways of annotating. In computer vision, many visual concepts such as region descriptions, objects,
attributes, and relationships, are annotated with bounding boxes, as shown in ~\cite{boundingboxannotation}. A common description of bounding boxes is \((x_c,y_c,w,h)\), where \((x_c, y_c)\) is the center location, \(w,h\) are the width and height of the bounding box, respectively.}

Objects without many orientations can be adequately annotated with this method. However, bounding boxes labeled in this way cannot accurately or compactly outline oriented instances such as text and objects in aerial images. In an extreme but actually common condition as shown in Fig.~\ref{fig:labeing-way}~(c) and (d), the overlap between two bounding boxes is so large that state-of-the-art object detection methods cannot differentiate them. In order to remedy this, we need to find an annotation method suitable for oriented objects.

\begin{figure*}[htb!]
\centering
\subfigure[]{
	\includegraphics[width=0.2\linewidth]{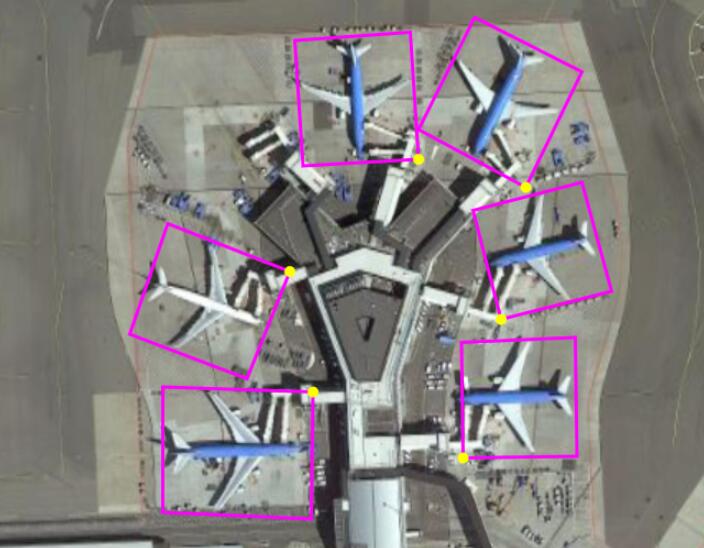}
}
\subfigure[]{
	\includegraphics[width=0.2\linewidth]{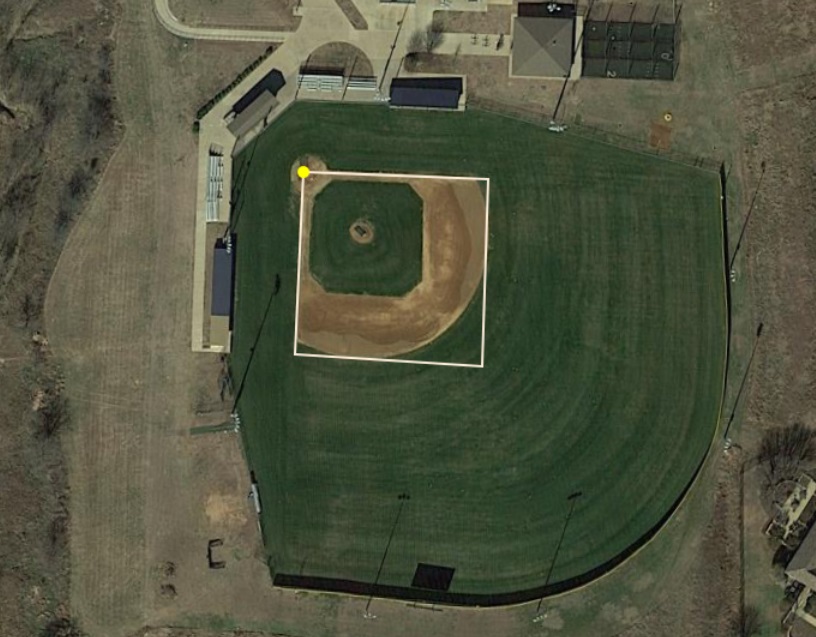}
}
\subfigure[]{
	\includegraphics[width=0.2\linewidth]{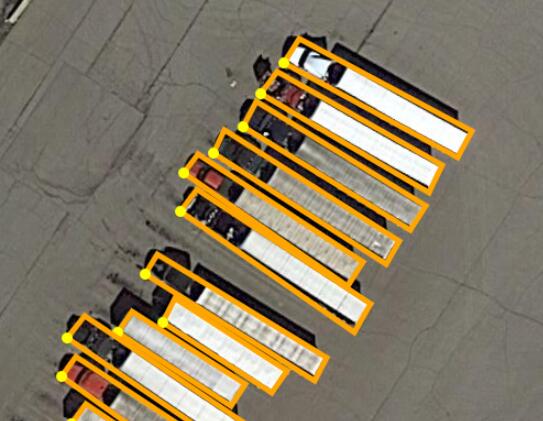}
}
\subfigure[]{
	\includegraphics[width=0.2\linewidth]{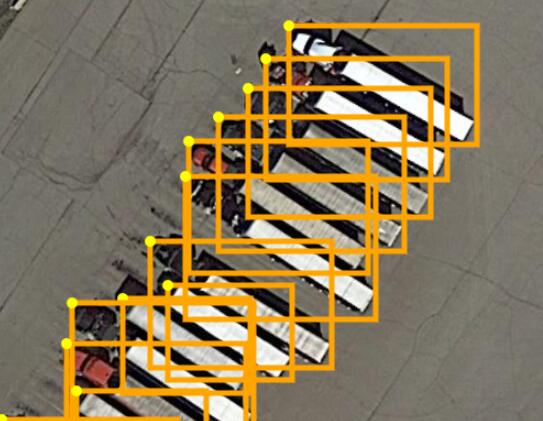}
}
\vspace{-3mm}
\caption{Visualization of adopted annotation method. The yellow point represents the starting point, which refers to:  (a) top left corner of a plane, (b) the center of sector-shaped baseball diamond, (c) top left corner of a large vehicle. (d) is a failure case of the horizontal rectangle annotation, which brings high overlap compared to (c).}
\label{fig:labeing-way}
\end{figure*}

{An option for annotating oriented objects is \(\theta\)-based oriented bounding box which is adopted in some text detection benchmarks ~\cite{MSRA}, namely \((x_c,y_c,w,h,\theta)\), where \(\theta\) denotes the angle from the horizontal direction of the standard bounding box. A flaw of this method is the inability to compactly enclose oriented objects with large deformation among different parts. Considering the complicated scenes and various orientations of objects in aerial images, we need to abandon this method and choose a more flexible and easy-to-understand way. An alternative is arbitrary quadrilateral bounding boxes, which can be denoted as \(\left \{ (x_i,y_i), i=1,2,3,4 \right \}\), where \((x_i, y_i)\) denotes the positions of the oriented bounding boxes' vertices in the image. The vertices are arranged in a clockwise order. This way is widely adopted in oriented scene text detection benchmarks~\cite{icdar15}. 
We draw inspiration from these researches and use arbitrary quadrilateral bounding boxes to annotate objects.}


{To make a more detailed annotation, as illustrated Fig.~\ref{fig:labeing-way}, we emphasize the importance of the first point \((x_1,y_1)\), which normally implies the ``head'' of the object. For helicopter, large vehicle, small vehicle, harbor, baseball diamond, ship and plane, we carefully denote their first point to enrich potential usages. While for soccer-ball field, swimming pool, bridge, ground track field, basketball court and tennis court, there are no visual clues to decide the first point, so we normally choose the top-left point as the starting point.}

Some samples of annotated patches (not the whole original image) in our dataset are shown in Fig.~\ref{fig:samples}.

It is worth noticing that, Papadopoulos \textit{et al.}~\cite{extremeclicking} have explored an alternative annotation method and verify its efficiency and robustness. We assume that the annotations would be more precise and robust with more elaborately designed annotation methods, and alternative annotation protocols would facilitate more efficient crowd-sourced image annotations.

\begin{figure*}[htb!]
\begin{center}
    \includegraphics[width=0.95\linewidth]{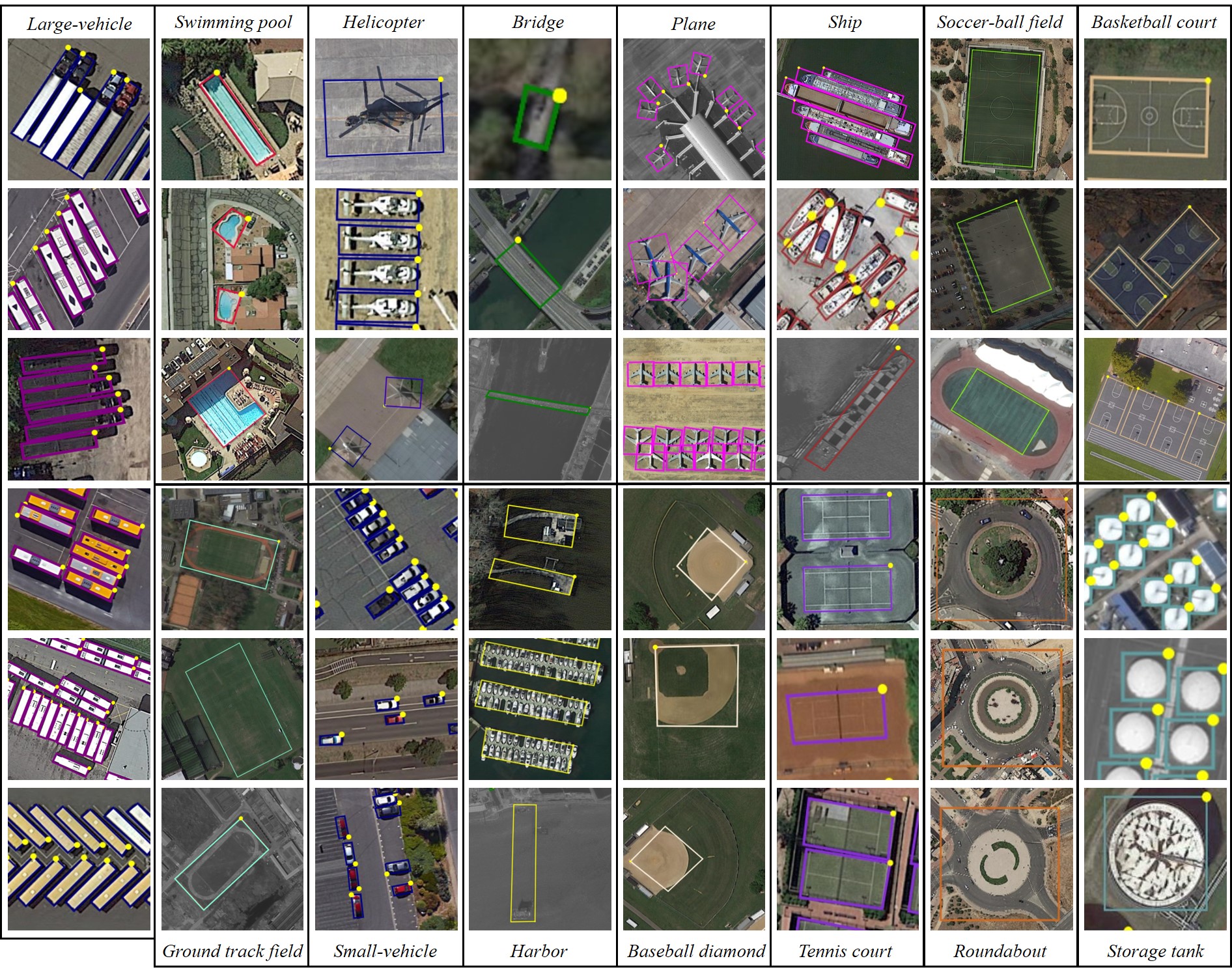}
\end{center}
\vspace{-3mm}
\caption{Samples of annotated images in DOTA. We show three samples per each category, except six for {\em large-vehicle}. }
\label{fig:samples}
\vspace{-2mm}
\end{figure*}

\subsection{Dataset splits}
In order to ensure that the training data and test data distributions approximately match, we randomly select half of the original images as the training set, 1/6 as validation set, and 1/3 as the testing set. We will publicly provide all the original images with ground truth for training set and validation set, but not for the testing set. For testing, we are currently building an evaluation server.

\section{Properties of DOTA}
\label{statistics_DOTA}

\subsection{Image size}
Aerial images are usually very large in size compared to those in natural images dataset. The original size of images in our dataset ranges from about $800 \times 800$ to about $ 4000 \times 4000$ while most images in regular datasets (e.g. PASCAL-VOC and MSCOCO) are no more than $1000 \times 1000$. 
We make annotations on the original full image without partitioning it into pieces to avoid the cases where a single instance is partitioned into different pieces.

\subsection{Various orientations of instances}
As shown in Fig.\ref{fig:large-example}~(f), our dataset achieves a good balance in the instances of different directions, which is significantly helpful for learning a robust detector. Moreover, our dataset is closer to real scenes because it is common to see objects in all kinds of orientations in the real world.

\begin{figure*}[htp!]
\begin{center}
 	\includegraphics[width=0.9\linewidth]{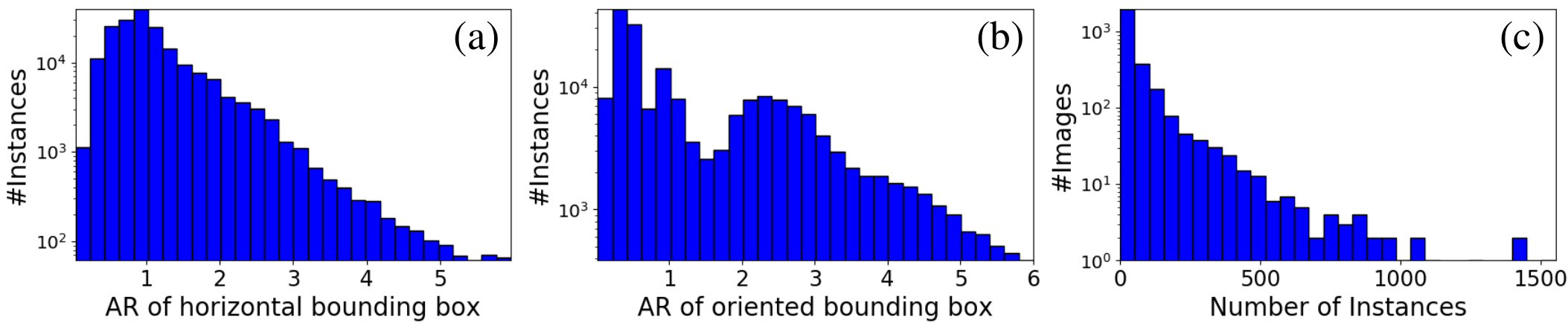}
\end{center}
\vspace{-3mm}
\caption{Statistics of instances in DOTA. AR denotes the aspect ratio. (a) The AR of horizontal bounding box. (b) The AR of oriented bounding box. (c) Histogram of number of annotated instances per image.}
\label{fig:ar_instances}
\end{figure*}

\subsection{Spatial resolution information}
We also provide the spatial resolution for each image in our dataset, which implies the actual size of an instance and plays a significant role in aerial object detection. The importance of spatial resolution for detection task are two folds. First, it allows the model to be more adaptive and robust for varieties of objects of the same category. It's known that objects appear smaller when seen from a distance. The same object with different sizes will trouble the model and hurt classification. However, a model can pay more attention to the shape with resolution information provided instead of objects' size. Second, it's better for fine-grained classification. For example, it will be simple to distinguish a small boat from a large warship.

Spatial resolution can also be used to filter mislabeled outliers in our dataset because intra-class varieties of actual sizes for most categories are limited. Outliers can be found by selecting the objects whose size is far different from those of the same category in a small range of spatial resolution.



\subsection{Various pixel size of categories}
Following the convention in~\cite{WIDERFACE}, we refer to the height of a horizontal bounding box, which we call pixel size for short, as a measurement for instance size. We divide all the instances in our dataset into three splits according to their height of horizontal bounding box: small for range from $10$ to $50$, middle for range from $50$ to $300$, and large for range above $300$.
Tab.~\ref{fig:size-distribution} illustrates the percentages of three instance splits in different datasets. It is clear that the PASCAL VOC dataset,  NWPU VHR-10 dataset and DLR 3K Munich Vehicle dataset are dominated by middle instances, middle instances and small instances, respectively. 
However, we achieve a good balance between small instances and middle instances, which is more similar to real-world scenes and thus, helpful to better capture different size of objects in practical applications.

It's worth noting that pixel size varies in different categories. 
For example, a vehicle may be as small as $30$, however, a bridge can be as large as $1200$, which is 40 times larger than a vehicle. The huge differences among instances from different categories make the detection task more challenging because models have to be flexible enough to handle extremely tiny and huge objects.
\begin{table}[htb!]
\footnotesize
\centering
\begin{tabular}{cccccccccccccccc}
\hline
Dataset & 10-50 pixel & 50-300 pixel & above 300 pixel \\ \hline
PASCAL VOC & 0.14 & 0.61 & 0.25 \\
MSCOCO & 0.43 & 0.49 & 0.08 \\ \hline
NWPU VHR-10 & 0.15 & 0.83 & 0.02 \\
DLR 3K Munich Vehicle  & 0.93 & 0.07 & 0 \\ 
\textbf{DOTA} & \textbf{0.57} & \textbf{0.41} & \textbf{0.02}\\
\hline 
\end{tabular}
\vspace{-2mm}
\caption{Comparison of instance size distribution of some datasets in aerial images and natural images.}
\vspace{-2mm}
\label{fig:size-distribution}
\end{table}

\subsection{Various aspect ratio of instances}
Aspect ratio is an essential factor for anchor-based models, such as Faster RCNN~\cite{FasterR-CNN} and YOLOv2~\cite{yolo9000}. 
We count two kinds of aspect ratio for all the instances in our dataset to provide a reference for better model design: 1) Aspect ratio of minimally circumscribed horizontal rectangle bounding box, 2) Aspect ratio of original quadrangle bounding box. Fig.~\ref{fig:ar_instances} illustrates these two types of distribution of aspect ratio for instances in our dataset. We can see that instances varies greatly in aspect ratio. Moreover, there are a large number of instances with a large aspect ratio in our dataset.

\subsection{Various instance density of images}
It is common for aerial images to contain thousands of instances, which is different from natural images. For example, images in ImageNet~\cite{Imagenet} contain on the average 2 categories and 2 instances, while MSCOCO contains 3.5 categories and 7.7 instances, respectively. 
Our dataset is much richer in instances per image, which can be up to 2000. Fig.~\ref{fig:ar_instances} illustrates the number of instances in our DOTA dataset.


With so many instances in a single image, it is unavoidable to see areas densely crowded with instances. For COCO, instances are not annotated one by one because occlusion makes it difficult to distinguish an instance from its neighboring  instances. In these cases, the group of instances is marked as one segment with attribute named ``crowd". However, this is not the case for aerial images because there are rarely occlusion due to the perspective from the above. Therefore, we can annotate all the instances in a dense area one by one. Fig.~\ref{fig:samples} shows examples of densely packed instances. Detecting objects in these cases poses an enormous challenge for the current detection methods.
\section{Evaluations}
We evaluate the state of the art object detection methods on DOTA. For horizontal object detection, we carefully select Faster R-CNN\footnote{\url{https://github.com/msracver/Deformable-ConvNets}}~\cite{FasterR-CNN}, R-FCN\footnote{\url{https://github.com/tensorflow/models/tree/master/research/object_detection}\label{web}}~\cite{R-FCN}, YOLOv2\footnote{\url{https://github.com/pjreddie/darknet}}~\cite{yolo9000} and SSD\footref{web}~\cite{SSD} as our benchmark testing algorithms for their excellent performance on general object detection. For oriented object detection, we modify the original Faster R-CNN algorithm such that it can predict properly oriented bounding boxes denoted as \(\left \{ (x_i,y_i), i=1,2,3,4 \right \}\).

Note that, the backbone networks are ResNet-101~\cite{resnet} for R-FCN and Faster R-CNN, InceptionV2~\cite{inception_v2} for SSD and customized GoogLeNet~\cite{Googlenet} for YOLOv2, respectively.

{\subsection{Tasks}
To comprehensively evaluate the state of the art deep learning based detection methods on DOTA, we propose two tasks, namely detection on \textit{horizontal bounding boxes} (\textbf{HBB} for short) and detection on \textit{oriented bounding boxes} (\textbf{OBB} for short). To be more specific, we evaluate those methods on two different kinds of ground truths, HBB or OBB, no matter how those methods were trained.}

\begin{table}[htb!]
\scriptsize
\centering
\begin{tabular}{c|c|c|c|c}
\hline
           & YOLOv2\cite{yolo9000}           & R-FCN\cite{R-FCN}          & FR-H\cite{FasterR-CNN}           & SSD\cite{SSD}         \\ \hline
    \textit{Plane}         & 76.9  & 81.01 & 80.32 & 57.85 \\
    \textit{BD}            & 33.87 & 58.96 & 77.55 & 32.79 \\
    \textit{Bridge}        & 22.73 & 31.64 & 32.86 & 16.14 \\
    \textit{GTF}           & 34.88 & 58.97 & 68.13 & 18.67 \\
    \textit{SV}            & 38.73 & 49.77 & 53.66 & 0.05 \\
    \textit{LV}            & 32.02 & 45.04 & 52.49 & 36.93 \\
    \textit{Ship}          & 52.37 & 49.29 & 50.04 & 24.74 \\
    \textit{TC}            & 61.65 & 68.99 & 90.41 & 81.16 \\
    \textit{BC}            & 48.54 & 52.07 & 75.05 & 25.1 \\
    \textit{ST}            & 33.91 & 67.42 & 59.59 & 47.47 \\
    \textit{SBF}           & 29.27 & 41.83 & 57    & 11.22 \\
    \textit{RA}            & 36.83 & 51.44 & 49.81 & 31.53 \\
    \textit{Harbor}        & 36.44 & 45.15 & 61.69 & 14.12 \\
    \textit{SP}            & 38.26 & 53.3  & 56.46 & 9.09 \\
    \textit{HC}            & 11.61 & 33.89 & 41.85 & 0 \\ \hline
    \textit{\textbf{Avg.}} & 39.2  & 52.58 & 60.46 & 29.86 \\ \hline     
\end{tabular}
\vspace{-2mm}
\caption{Numerical results (AP) of baseline models evaluated with HBB ground truths. The short names for categories are defined as: {\em BD--Baseball diamond, GTF--Ground field track, SV--Small vehicle,  LV--Large vehicle, TC--Tennis court, BC--Basketball court, SC--Storage tank, SBF--Soccer-ball field, RA--Roundabout, SP--Swimming pool, and  HC--Helicopter}. FR-H means \textbf{F}aster \textbf{R}-CNN~\cite{FasterR-CNN} trained on \textbf{H}orizontal bounding boxes. FR-O means \textbf{F}aster \textbf{R}-CNN~\cite{FasterR-CNN} trained on \textbf{O}riented bounding boxes.}
\label{numericalresults}
\end{table}

\begin{table}[htb!]
  \scriptsize
  \centering
\begin{tabular}{c|c|c|c|c||c}
\hline
                             &  YOLOv2~\cite{yolo9000}     &   R-FCN~\cite{R-FCN}         & SSD~\cite{SSD}     & FR-H~\cite{FasterR-CNN}  & FR-O   \\ \hline

\textit{Plane}                    &    52.75 & 39.57 & 41.06 & 49.74 & 79.42 \\
\textit{BD}                   &     24.24 & 46.13 & 24.31 & 64.22 & 77.13 \\
\textit{Bridge}                   &     10.6  & 3.03  & 4.55  & 9.38  & 17.7 \\
\textit{GTF}                   &     35.5  & 38.46 & 17.1  & 56.66 & 64.05 \\
\textit{SV}                   &     14.36 & 9.1   & 15.93 & 19.18 & 35.3 \\
\textit{LV}                   &     2.41  & 3.66  & 7.72  & 14.17 & 38.02 \\
\textit{Ship}                   &     7.37  & 7.45  & 13.21 & 9.51  & 37.16 \\
\textit{TC}                   &     51.79 & 41.97 & 39.96 & 61.61 & 89.41 \\
\textit{BC}                   &     43.98 & 50.43 & 12.05 & 65.47 & 69.64 \\
\textit{ST}                   &     31.35 & 66.98 & 46.88 & 57.52 & 59.28 \\
\textit{SBF}                   &     22.3  & 40.34 & 9.09  & 51.36 & 50.3 \\
\textit{RA}                   &     36.68 & 51.28 & 30.82 & 49.41 & 52.91 \\
\textit{Harbor}                   &     14.61 & 11.14 & 1.36  & 20.8  & 47.89 \\
\textit{SP}                   &     22.55 & 35.59 & 3.5   & 45.84 & 47.4 \\
\textit{HC}                   &     11.89 & 17.45 & 0     & 24.38 & 46.3 \\ \hline
\textit{\textbf{Avg.}}                   &     25.492 & 30.84 & 17.84 & 39.95 & 54.13 \\ \hline   \end{tabular}
  \caption{Numerical results (AP) of baseline models evaluated with OBB ground truths. The short names are defined the same as depicted in Tab.~\ref{numericalresults}. Note that only FR-O~\cite{FasterR-CNN} is trained with OBB.}
  \label{OBBresults}%
\vspace{-3mm}
\end{table}%

\subsection{Evaluation prototypes}

Images in DOTA are so large that they cannot be directly sent to CNN-based detectors. Therefore, we crop a series of \(1024\times1024\) patches from the original images with a stride set to 512. Note that some complete objects may be cut into two parts during the cropping process. For convenience, we denote the area of the original object as \(A_o\), and the area of divided parts \(P_i,\, (i=1,2)\) as \(a_i,\,(i=1,2)\). Then we compute the parts areas over the original object area: 
\[U_{i}=\frac{a_{i}}{A_{o}}.\]
Finally, we label the part \(P_i\) with \(U_{i}< 0.7\) as \textit{difficult} and for the other one, we keep it the same as the original annotation. For the vertices of the newly generated parts, we need to ensure they can be described as an oriented bounding box with 4 vertices in the clockwise order with a fitting method.

In the testing phase, first we send the cropped image patches to obtain temporary results and then we combine the results together to restore the detecting results on the original image. Finally, we use {\em non-maximum suppression} (NMS) on these results based on the predicted classes. We keep the threshold of NMS as $0.3$ for the HBB experiments and $0.1$ for the oriented experiments. In this way, we indirectly  train and test CNN-based models on DOTA. 

For evaluation metrics, we adopt the same mAP calculation as for PASCAL VOC.

\subsection{Baselines with horizontal bounding boxes}
\label{sec:horizontal_baseline}

Ground truths for HBB experiments are generated by calculating the axis-aligned bounding boxes over original annotated bounding boxes. To make it fair, we keep all the experiments' settings and hyper parameters the same as depicted in corresponding papers~\cite{FasterR-CNN,R-FCN,yolo9000,SSD}.




The experimental results of HBB prediction are shown in Tab.~\ref{numericalresults}. Note that results of SSD is much lower than other models. We suspect it should be attributed to the random crop operation in SSD's data augmentation strategies, which is quite useful in general object detection while degrades in aerial object detection for tremendous small training instances. The results further indicate the huge differences between aerial and general objects with respect to instance sizes.

\subsection{Baselines with oriented bounding boxes}

Prediction of OBB is difficult because the state of the art detection methods are not designed for oriented objects. Therefore, we choose Faster R-CNN as the base framework for its accuracy and efficiency and then modify it to predict oriented bounding boxes. 

RoIs (Region of Interests) generated by RPN (Region Proposal Network) are rectangles which can be written as \(R=(x_{min}, y_{min}, x_{max}, y_{max})\), for a more detailed interpretation, \(R=\left \{(x_i, y_i),i=1,2,3,4\right \}\), where \(x_1=x_4=x_{min}, x_2=x_3=x_{max}, y_1=y_2=y_{min}, y_3=y_4=y_{max}\). In R-CNN procedure, each RoI is attached to a ground truth oriented bounding box written as \(G = \left \{(g_{xi}, g_{yi}),i=1,2,3,4\right \}\). Then R-CNN's output target  \(T = \left \{(t_{xi}, t_{yi}),i=1,2,3,4\right \}\) is calculated by following equations,
\begin{align}
t_{xi} &=(g_{xi}-x_i)/w,  \\
t_{yi} &=(g_{yi}-y_i)/h,
\end{align}
where $i=1,2,3,4$, \(w=x_{max}-x_{min}\), and \(h=y_{max}-y_{min}\).

Other settings and hyper parameters are kept the same as depicted in Faster R-CNN~\cite{FasterR-CNN}. The numerical results are shown in Tab.~\ref{OBBresults}.
{To make a comparison to our implemented Faster R-CNN for OBB, we evaluate YOLOv2, R-FCN, SSD and Faster R-CNN trained on HBB with the OBB ground truth. As shown in Tab.\ref{OBBresults}, the results of those methods trained on HBB are much lower than Faster R-CNN trained on OBB, indicating that for oriented object detection in aerial scenes, those methods should be adjusted accordingly.}

\subsection{Experimental analysis}
When analyzing the results exhibited in Table.~\ref{numericalresults}, performances in categories like small vehicle, large vehicle and ship are far from satisfactory, which attributes to their small size and densely crowded locations in aerial images. As a contrast, large and discrete objects, like planes, swimming pools and tennis courts, the performances are rather fair. 

In Fig.~\ref{fig:exp_H_O}, we compare the results between object detection experiments of HBB and OBB. For densely packed and oriented objects shown in Fig.~\ref{fig:exp_H_O}~(a) and (b), location precision of objects in HBB experiments are much lower than OBB experiments and many results are suppressed through post-progress operations. So OBB regression is the correct way for oriented object detection that can be really integrated to real applications. In Fig.~\ref{fig:exp_H_O}~(c), large aspect ratio objects annotated in OBB style like (harbor, bridge) are hard for current detectors to regress.  But in HBB style, those objects usually have normal aspect ratios and as a consequence, results seem to be fairly good as shown in Fig.~\ref{fig:exp_H_O}~(d). However in extremely dense scenes, e.g in Fig.~\ref{fig:exp_H_O}~(e) and (f), results of HBB and OBB are all not satisfying which implies the defects of current detectors.


\begin{figure}[htb!]
\begin{center}
\includegraphics[width=0.99\linewidth]{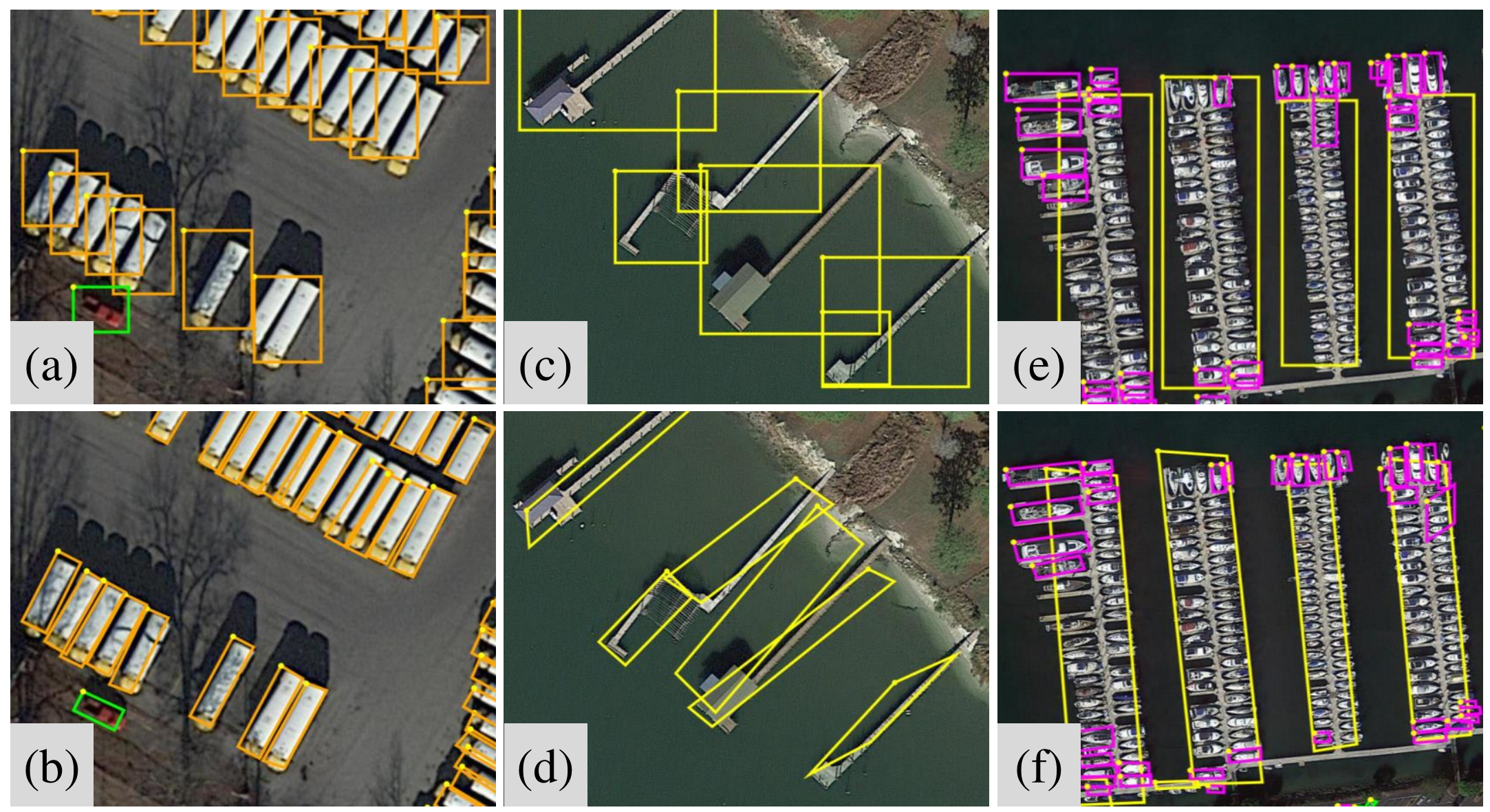}
\end{center}
\caption{Visualization results of testing on DOTA using well-trained Faster R-CNN. \textbf{TOP} and \textbf{Bottom} respectively illustrate the results for HBB and OBB in cases of orientation, large aspect ratio, and density.}
\label{fig:exp_H_O}
\end{figure}

\section{Cross-dataset validations}
The cross dataset generalization~\cite{torralba2011unbiased} is an  evaluation for the generalization ability of a dataset. 
We choose the UCAS-AOD dataset~\cite{ucas-aod} to do cross-dataset generalization for its comparatively large number of data comparing to other aerial object detection datasets. 
For there are no official data splits for UCAS-AOD, we randomly select 1110 for training and 400 for testing.
We choose YOLOv2 as the testing detector for all experiments described below and HBB-style annotations for all ground truths. Input image size is changed to $960 \times 544$ around the original image sizes in UCAS-AOD while other setting kept unchanged.

Results are shown in Tab.~\ref{fig:cross-dataset}.
The performance difference across two datasets is 35.8 for YOLOv2-A and 15.6 for YOLOv2-D models, respectively. It suggests that DOTA hugely covers UCAS-AOD and furthermore has more patterns and properties that are not shared in UCAS-AOD. And both models get a low results on DOTA which reflects that DOTA is much more challenging.



\begin{table}[htb!]
\small
\centering
\begin{tabular}{ccccc}
\hline
Testing set                  & Detector    & Plane & Small-vehicle & Avg.  \\ \hline
\multirow{2}{*}{UCAS-AOD} & YOLOv2-A & 90.66 & 88.17         & 89.41 \\ \cline{2-5} 
                          & YOLOv2-D & 87.18 & 65.13         & 76.15 \\ \hline
\multirow{2}{*}{DOTA}     & YOLOv2-A & 62.92 & 44.17         & 53.55 \\ \cline{2-5} 
                          & YOLOv2-D & 74.83 & 46.18         & 60.51 \\ \hline
\end{tabular}
\vspace{-2mm}
\caption{Results of cross-dataset generalization. \textbf{Top:} Detection performance evaluated on \textbf{UCAS-AOD}. \textbf{Bottom:} Detection performance evaluated on \textbf{DOTA}. YOLOv2-A and YOLOv2-D are trained with UCAS-AOD and DOTA, respectively.}
\label{fig:cross-dataset}
\end{table}


\section{Conclusion}
We build a large-scale dataset for oriented objects detection in aerial images which is much larger than any existing datasets in this field. In contrast to general object detection benchmarks, we annotate a huge number of well-distributed oriented objects with oriented bounding boxes. We assume this dataset is challenging but very similar to natural aerial scenes, which are more appropriate for practical applications. We also establish a benchmark for object detection in aerial images and show the feasibility to produce oriented bounding boxes by modifying a mainstream detection algorithm. 

Detecting densely packed small instances and extremely large instances with arbitrary orientations in a large picture would be particularly meaningful and challenging. 
We believe DOTA will not only promote the development of object detection algorithms in Earth Vision, but also pose interesting algorithmic questions to general object detection in computer vision.

\section{Acknowledgement}
We thank Fan Hu, Pu Jin, Xinyi Tong, Xuan Hu, Zhipeng Dong, Liang Wu, Jun Tang, Linyan Cui, Duoyou Zhou, Tengteng Huang, and all the others who involved in the annotations of DOTA.

{\small
\bibliographystyle{ieee}
\bibliography{aerial_detection_dataset}
}

\end{document}